\documentclass{article}
\usepackage{spconf,amsmath,graphicx}

\usepackage{tikz}
\usetikzlibrary{arrows}
\usepackage{amsfonts}
\usepackage{hyperref}

\newcommand{\mycomment}[1]{}

\title{EXTRACTER: Efficient Texture Matching with Attention and Gradient Enhancing for Large Scale Image Super Resolution}
\name{Esteban Reyes-Saldaña, Mariano Rivera\thanks{\noindent Code available at \url{ https://github.com/esteban-rs/EXTRACTER}. \\ This work has been submitted to the IEEE for possible publication. Copyright may be transferred without notice, after which this version may no longer be accessible.}}
\address{Centro de Investigacion en Matematicas A.C.\\
        Guanajuato, Gto., 36023 Mexico \\
        \{esteban.reyes, mrivera\}@cimat.mx
}
\begin{document}
%
\maketitle
\begin{abstract}
Recent Reference-Based image super-resolution (RefSR) has improved SOTA deep methods introducing attention mechanisms to enhance low-resolution images by transferring high-resolution textures from a reference high-resolution image. The main idea is to search for matches between patches using LR and Reference image pair in a feature space and merge them using deep architectures. However, existing methods lack the accurate search of textures. They divide images into as many patches as possible, resulting in inefficient memory usage, and cannot manage large images. Herein, we propose a deep search with a more efficient memory usage that reduces significantly the number of image patches and finds the $k$ most relevant texture match for each low-resolution patch over the high-resolution reference patches, resulting in an accu\-rate texture match. We enhance the Super Resolution result adding gradient density information using a simple residual architecture showing competitive metrics results: PSNR and SSMI.
\end{abstract}
\section{Introduction}
\label{sec:intro}

\mycomment{
There exist two paradigms for Image Super-Resolution: Single Image (SISR) and Reference Based Super Resolution (RefSR). SISR suffers from really blurred reconstructed images (as expected since the low-resolution input image is missing some important details). 
}
The paradigm Image Reference-Super Resolution aims to recover high-resolution Images by transferring accurate textures from a reference image (with a centrain similarity degree) reducing burred and artifacts. In recent years, vision transformers have improved super-resolution results.  For example, TTSR\cite{TTSR} introduces attention to Ref-Super Resolution by successfully transferring textures from the Ref image. They use a learnable VGG pre-trained feature extractor to obtain attention matrices $ Q, K, V)$ to perform a cross-attention mechanism to find the best features for the SR reconstruction. 

Lin et al. \cite{Dual_Projection_Fusion} proposed a novel low-resolution backbone capable of extracting a best feature representation and adding a branch to refine the low-resolution and reference features. Some other works \cite{CIMR, Matter} claim that a better texture search is required in order to obtain less blurred images and use multiple reference images for a more accurate pattern search. Gou et al. \cite{doubleLayer} enhance memory efficiency by using low-resolution dimensions to find correlations and filtering patch matches for enhancing the final result and adding gradient information using a pre-existing SISR model for the final result.

To address the above problems, we propose a search stra-tegy to efficiently split the images into patches, find the $top_k$ HR matches for each LR patch, and add structural information for enhancing the Super-Resolution result. Specifically, we first extract deep features from a VGG19-based architecture. Different from \cite{TTSR} and most of the recent methods, we split images into patches using a $ 6 \times 6 $ window (instead of $3\times 3$) for the deepest feature level, resulting in a more memory efficient usage that can allow us to use large-scale images. Second, we propose a research strategy but different from \cite{doubleLayer}, we use $top_k$ matches between the low-resolution and ref patches instead of the max feature for each low-resolution patch. Finally, we merge textures at different scales and add gradient density information form a better spatial reconstruction using a simple residual network.

The primary contributions of this paper are. First. we introduce a Search and Transfer module to identify correlations between low-resolution and reference patches; we use larger window in  with state-of-the-art (SOTA) methods. This significantly reduces the dimensionality of the correlation matrix and allows to use the top-$k$ matches to enhance texture transfer. Second, we introduce a Gradient Density-Enhancing Module (GDE) to improve the merging of textures from diffe\-rent deep levels while considering gradient density information. This module is implemented by a straightforward recu\-rrent network. And third, we conduct extensive experiments on benchmark datasets that provide us strong evidence that the proposal overcomes SOTA methods.

\section{Related Work}


In recent years, Single Image Super Resolution (SISR) improved super-resolution methods by using residual blocks\cite{grad2} and designing deeper networks. These methods use $\mathcal{L}_1$ and $\mathcal{L}_2$ losses as the training objective functions that have demonstrated nonaccuracy for human perception \cite{TTSR}. To solve this, novel methods use a GAN strategy\cite{srgan} resulting in better satisfying results or adopt classic computer vision transformation such as gradient mapping \cite{grad1}.

Since the appearance of vision transformers, vision tasks has been improved.  For example, TTSR\cite{TTSR} introduces cross-attention to Ref-Super Resolution for transferring textures: a patch matching based technique robust to miss-alignment problems \cite{pixel,crossnet}. Based on TTSR,  Lin et al. \cite{Dual_Projection_Fusion} add channel-wise attention. \cite{CIMR, Matter} and use multiple image patches for transferring textures, resulting in better results. In this direction, cross-attention mechanisms are used and better memory usage is required. Gou et al. \cite{doubleLayer} enhance memory efficiency by using low-resolution dimensions to find correlations and use classical vision transformation for structural reconstruction, such as gradient density flow.

\vspace{-3mm}
\section{Method}
\vspace{-1mm}

\begin{figure}
    \centering
    \definecolor{ffzzqq}{rgb}{1,0.6,0}
\definecolor{ffccww}{rgb}{1,0.8,0.4}
\definecolor{ccwwff}{rgb}{0.8,0.4,1}
\definecolor{cczzff}{rgb}{0.8,0.6,1}
\definecolor{dqfqcq}{rgb}{0.8156862745098039,0.9411764705882353,0.7529411764705882}
\definecolor{ccccff}{rgb}{0.8,0.8,1}
\definecolor{ffzztt}{rgb}{1,0.6,0.2}
\definecolor{afeeee}{rgb}{0.6862745098039216,0.9333333333333333,0.9333333333333333}
\begin{tikzpicture}[line cap=round,line join=round,>=triangle 45,x=0.35cm,y=0.35cm]
\clip(1.9,-4.1) rectangle (26.1,10);
\fill[line width=0.5pt,color=afeeee,fill=afeeee,fill opacity=1] (2,2) -- (2,0) -- (4,0) -- (4,2) -- cycle;
\fill[line width=0.5pt,color=afeeee,fill=afeeee,fill opacity=1] (2,3) -- (4,3) -- (4,5) -- (2,5) -- cycle;
\fill[line width=0.5pt,color=afeeee,fill=afeeee,fill opacity=1] (1.9679909899501222,8) -- (1.9679909899501222,6) -- (3.9679909899501222,6) -- (3.9679909899501222,8) -- cycle;
\fill[line width=0.5pt,color=afeeee,fill=afeeee,fill opacity=1] (3,10) -- (3,9) -- (4,9) -- (4,10) -- cycle;
\fill[line width=0.5pt,color=ffzztt,fill=ffzztt,fill opacity=1] (7,10) -- (7,9) -- (8,9) -- (8,10) -- cycle;
\fill[line width=0.5pt,color=ccccff,fill=ccccff,fill opacity=1] (11,1) -- (11,0) -- (12,0) -- (12,1) -- cycle;
\fill[line width=0.5pt,color=dqfqcq,fill=dqfqcq,fill opacity=1] (11,7) -- (11,6) -- (12,6) -- (12,7) -- cycle;
\fill[line width=0.5pt,color=dqfqcq,fill=dqfqcq,fill opacity=1] (11.206272092959187,7) -- (11.2,7.2) -- (12.197314548032221,7.197314548032221) -- (12.200153666666512,6.2426628848614865) -- (12,6.23994341515434) -- (12,7) -- cycle;
\fill[line width=0.5pt,color=dqfqcq,fill=dqfqcq,fill opacity=1] (11.399462909606445,7.199462909606444) -- (11.397314548032222,7.397314548032221) -- (12.397314548032222,7.397314548032221) -- (12.398978368326667,6.637980342440351) -- (12.198980918189013,6.636998743533574) -- (12.197314548032221,7.197314548032221) -- cycle;
\fill[line width=0.5pt,color=dqfqcq,fill=dqfqcq,fill opacity=1] (11,3) -- (11,1.5) -- (12.5,1.5) -- (12.5,3) -- cycle;
\fill[line width=0.5pt,color=dqfqcq,fill=dqfqcq,fill opacity=1] (11,5) -- (11,3.8) -- (12.2,3.8) -- (12.2,5) -- cycle;
\fill[line width=0.5pt,color=dqfqcq,fill=dqfqcq,fill opacity=1] (11.2,5) -- (11.2,5.2) -- (12.4,5.2) -- (12.4,4) -- (12.2,4) -- (12.2,5) -- cycle;
\fill[line width=0.5pt,color=dqfqcq,fill=dqfqcq,fill opacity=1] (11.399832602781277,5.2) -- (11.4,5.4) -- (12.6,5.4) -- (12.6,4.2) -- (12.4,4.2) -- (12.4,5.2) -- cycle;
\fill[line width=0.5pt,color=dqfqcq,fill=dqfqcq,fill opacity=1] (11.2,3) -- (11.2,3.2) -- (12.7140650234447,3.1966792551308933) -- (12.719602158509842,1.7129950246301735) -- (12.5,1.7104975587538556) -- (12.5,3) -- cycle;
\fill[line width=0.5pt,color=dqfqcq,fill=dqfqcq,fill opacity=1] (11.402839803366632,3.199555118686549) -- (11.390954914337485,3.4994959422876737) -- (12.910354342952765,3.5007778993086807) -- (12.924394360367906,1.902802431230326) -- (12.718892746154541,1.9030832041849488) -- (12.7140650234447,3.1966792551308933) -- cycle;
\fill[line width=0.5pt,color=cczzff,fill=cczzff,fill opacity=1] (11.2,1) -- (11.2,1.2) -- (12.2,1.2) -- (12.2,0.2) -- (12,0.2) -- (12,1) -- cycle;
\fill[line width=0.5pt,color=ccwwff,fill=ccwwff,fill opacity=1] (11.4,1.2) -- (11.4,1.4) -- (12.4,1.4) -- (12.4,0.4) -- (12.2,0.4) -- (12.2,1.2) -- cycle;
\fill[line width=0.5pt,color=ffccww,fill=ffccww,fill opacity=1] (16,8) -- (16,4) -- (26,4) -- (26,8) -- cycle;
\fill[line width=0.5pt,color=dqfqcq,fill=dqfqcq,fill opacity=1] (16,3) -- (16,1) -- (18,1) -- (18,3) -- cycle;
\fill[line width=0.5pt,color=dqfqcq,fill=dqfqcq,fill opacity=1] (19.710992720268464,3) -- (19.710992720268464,1) -- (21.710992720268464,1) -- (21.710992720268464,3) -- cycle;
\fill[line width=0.5pt,color=dqfqcq,fill=dqfqcq,fill opacity=1] (22.5,2.5) -- (22.5,1) -- (24,1) -- (24,2.5) -- cycle;
\fill[line width=0.5pt,color=dqfqcq,fill=dqfqcq,fill opacity=1] (24.702232277835563,2.2106277501362612) -- (24.692645527201645,1.0183471930668224) -- (25.992645527201645,1.0183471930668224) -- (25.992645527201645,2.2183471930668226) -- cycle;

\fill[line width=0.5pt,color=ffccww,fill=ffccww,fill opacity=1] (16,-1) -- (16,-4) -- (22,-4) -- (22,-1) -- cycle;
\fill[line width=0.5pt,color=afeeee,fill=afeeee,fill opacity=1] (24,-2) -- (24,-4) -- (26,-4) -- (26,-2) -- cycle;

\fill[line width=0.5pt,color=ffzzqq,fill=ffzzqq,fill opacity=1] (7,-2) -- (7,-3) -- (8,-3) -- (8,-2) -- cycle;
\fill[line width=0.5pt,color=dqfqcq,fill=dqfqcq,fill opacity=1] (4.5,-2) -- (4.5,-3) -- (5.5,-3) -- (5.5,-2) -- cycle;
\fill[line width=0.5pt,color=ffccww,fill=ffccww,fill opacity=1] (6,7.5) -- (6,0) -- (10,2) -- (10,6) -- cycle;
\draw [line width=0.5pt] (2,2)-- (2,0);
\draw [line width=0.5pt] (2,0)-- (4,0);
\draw [line width=0.5pt] (4,0)-- (4,2);
\draw [line width=0.5pt] (2,3)-- (4,3);
\draw [line width=0.5pt] (4,3)-- (4,5);
\draw [line width=0.5pt] (4,5)-- (2,5);
\draw [line width=0.5pt] (2,5)-- (2,3);
\draw [line width=0.5pt] (1.9679909899501222,8)-- (1.9679909899501222,6);
\draw [line width=0.5pt] (1.9679909899501222,6)-- (3.9679909899501222,6);
\draw [line width=0.5pt] (3.9679909899501222,6)-- (3.9679909899501222,8);
\draw [line width=0.5pt] (3.9679909899501222,8)-- (1.9679909899501222,8);
\draw [line width=0.5pt] (3,10)-- (3,9);
\draw [line width=0.5pt] (3,9)-- (4,9);
\draw [line width=0.5pt] (4,9)-- (4,10);
\draw [line width=0.5pt] (4,10)-- (3,10);
\draw [-latex] (3.9679909899501222,6.5) -- (6,6.5);
\draw [line width=0.5pt] (7,10)-- (7,9);
\draw [line width=0.5pt] (7,9)-- (8,9);
\draw [line width=0.5pt] (8,9)-- (8,10);
\draw [line width=0.5pt] (8,10)-- (7,10);
\draw [-latex] (4,9.590958269541302) -- (7,9.60569720076159);
\draw [line width=0.5pt] (11,1)-- (11,0);
\draw [line width=0.5pt] (11,0)-- (12,0);
\draw [line width=0.5pt] (12,0)-- (12,1);
\draw [line width=0.5pt] (12,1)-- (11,1);
\draw [-latex] (4,4) -- (6,4);
\draw [-latex] (4,1) -- (6,1);
\draw [line width=0.5pt] (11,7)-- (11,6);
\draw [line width=0.5pt] (11,6)-- (12,6);
\draw [line width=0.5pt] (12,6)-- (12,7);
\draw [line width=0.5pt] (12,7)-- (11,7);
\draw [line width=0.5pt] (11.206272092959187,7)-- (11.2,7.2);
\draw [line width=0.5pt] (11.2,7.2)-- (12.197314548032221,7.197314548032221);
\draw [line width=0.5pt] (12.197314548032221,7.197314548032221)-- (12.200153666666512,6.2426628848614865);
\draw [line width=0.5pt] (12.200153666666512,6.2426628848614865)-- (12,6.23994341515434);
\draw [line width=0.5pt,color=dqfqcq] (12,6.23994341515434)-- (12,7);
\draw [line width=0.5pt,color=dqfqcq] (12,7)-- (11.206272092959187,7);
\draw [line width=0.5pt] (11.399462909606445,7.199462909606444)-- (11.397314548032222,7.397314548032221);
\draw [line width=0.5pt] (11.397314548032222,7.397314548032221)-- (12.397314548032222,7.397314548032221);
\draw [line width=0.5pt] (12.397314548032222,7.397314548032221)-- (12.398978368326667,6.637980342440351);
\draw [line width=0.5pt] (12.398978368326667,6.637980342440351)-- (12.198980918189013,6.636998743533574);
\draw [line width=0.5pt,color=dqfqcq] (12.198980918189013,6.636998743533574)-- (12.197314548032221,7.197314548032221);
\draw [line width=0.5pt,color=dqfqcq] (12.197314548032221,7.197314548032221)-- (11.399462909606445,7.199462909606444);
\draw [line width=0.5pt] (11,3)-- (11,1.5);
\draw [line width=0.5pt] (11,1.5)-- (12.5,1.5);
\draw [line width=0.5pt] (12.5,1.5)-- (12.5,3);
\draw [line width=0.5pt] (12.5,3)-- (11,3);
\draw [line width=0.5pt] (11,5)-- (11,3.8);
\draw [line width=0.5pt] (11,3.8)-- (12.2,3.8);
\draw [line width=0.5pt] (12.2,3.8)-- (12.2,5);
\draw [line width=0.5pt] (12.2,5)-- (11,5);
\draw [line width=0.5pt] (11.2,5)-- (11.2,5.2);
\draw [line width=0.5pt] (11.2,5.2)-- (12.4,5.2);
\draw [line width=0.5pt] (12.4,5.2)-- (12.4,4);
\draw [line width=0.5pt] (12.4,4)-- (12.2,4);
\draw [line width=0.5pt] (12.2,4)-- (12.2,5);
\draw [line width=0.5pt] (12.2,5)-- (11.2,5);
\draw [line width=0.5pt] (11.399832602781277,5.2)-- (11.4,5.4);
\draw [line width=0.5pt] (11.4,5.4)-- (12.6,5.4);
\draw [line width=0.5pt] (12.6,5.4)-- (12.6,4.2);
\draw [line width=0.5pt] (12.6,4.2)-- (12.4,4.2);
\draw [line width=0.5pt] (12.4,4.2)-- (12.4,5.2);
\draw [line width=0.5pt] (12.4,5.2)-- (11.399832602781277,5.2);
\draw [line width=0.5pt] (11.2,7.2)-- (12.197314548032226,7.197314548032214);
\draw [line width=0.5pt] (12.197314548032226,7.197314548032214)-- (12.200153666666518,6.242662884861544);
\draw [line width=0.5pt] (12,7)-- (12,6.239943415154397);
\draw [line width=0.5pt] (12,7)-- (11,7);
\draw [line width=0.5pt] (11.2,3)-- (11.2,3.2);
\draw [line width=0.5pt] (11.2,3.2)-- (12.7140650234447,3.1966792551308933);
\draw [line width=0.5pt] (12.7140650234447,3.1966792551308933)-- (12.719602158509842,1.7129950246301735);
\draw [line width=0.5pt] (12.719602158509842,1.7129950246301735)-- (12.5,1.7104975587538556);
\draw [line width=0.5pt] (12.5,1.7104975587538556)-- (12.5,3);
\draw [line width=0.5pt] (12.5,3)-- (11.2,3);
\draw [line width=0.5pt] (11.402839803366632,3.199555118686549)-- (11.390954914337485,3.4994959422876737);
\draw [line width=0.5pt] (11.390954914337485,3.4994959422876737)-- (12.910354342952765,3.5007778993086807);
\draw [line width=0.5pt] (12.910354342952765,3.5007778993086807)-- (12.924394360367906,1.902802431230326);
\draw [line width=0.5pt] (12.924394360367906,1.902802431230326)-- (12.718892746154541,1.9030832041849488);
\draw [line width=0.5pt] (12.718892746154541,1.9030832041849488)-- (12.7140650234447,3.1966792551308933);
\draw [line width=0.5pt] (12.7140650234447,3.1966792551308933)-- (11.402839803366632,3.199555118686549);
\draw [line width=0.5pt] (11.2,1)-- (11.2,1.2);
\draw [line width=0.5pt] (11.2,1.2)-- (12.2,1.2);
\draw [line width=0.5pt] (12.2,1.2)-- (12.2,0.2);
\draw [line width=0.5pt] (12.2,0.2)-- (12,0.2);
\draw [line width=0.5pt] (12,0.2)-- (12,1);
\draw [line width=0.5pt] (12,1)-- (11.2,1);
\draw [line width=0.5pt] (11.4,1.2)-- (11.4,1.4);
\draw [line width=0.5pt] (11.4,1.4)-- (12.4,1.4);
\draw [line width=0.5pt] (12.4,1.4)-- (12.4,0.4);
\draw [line width=0.5pt] (12.4,0.4)-- (12.2,0.4);
\draw [line width=0.5pt] (12.2,0.4)-- (12.2,1.2);
\draw [line width=0.5pt] (12.2,1.2)-- (11.4,1.2);
\draw [line width=0.5pt] (16,8)-- (16,4);
\draw [line width=0.5pt] (16,4)-- (26,4);
\draw [line width=0.5pt] (26,4)-- (26,8);
\draw [line width=0.5pt] (26,8)-- (16,8);
\draw [-latex] (12.39820249930647,6.992071399066152) -- (16,6.968436602762954);
\draw [-latex] (12.6,4.5) -- (16,6.376644234649023);
\draw [-latex] (12.914116633353178,3) -- (16,5.595478308738634);
\draw [-latex] (12.4,0.8) -- (16,4.60126713030723);
\draw [line width=0.5pt] (2,2)-- (4,2);
\draw [line width=0.5pt] (10,5)-- (10,6.5);
\draw [line width=0.5pt] (10,2)-- (10,0.5);
\draw [-latex] (10,6.5) -- (11,6.481827777355196);
\draw [-latex] (10,0.5) -- (11,0.4968281068260615);
\draw [line width=0.5pt] (16,3)-- (16,1);
\draw [line width=0.5pt] (16,1)-- (18,1);
\draw [line width=0.5pt] (18,1)-- (18,3);
\draw [line width=0.5pt] (18,3)-- (16,3);
\draw [line width=0.5pt] (19.710992720268464,3)-- (19.710992720268464,1);
\draw [line width=0.5pt] (19.710992720268464,1)-- (21.710992720268464,1);
\draw [line width=0.5pt] (21.710992720268464,1)-- (21.710992720268464,3);
\draw [line width=0.5pt] (21.710992720268464,3)-- (19.710992720268464,3);
\draw [line width=0.5pt] (22.5,2.5)-- (22.5,1);
\draw [line width=0.5pt] (22.5,1)-- (24,1);
\draw [line width=0.5pt] (24,1)-- (24,2.5);
\draw [line width=0.5pt] (24,2.5)-- (22.5,2.5);
\draw [line width=0.5pt] (24.702232277835563,2.2106277501362612)-- (24.692645527201645,1.0183471930668224);
\draw [line width=0.5pt] (24.692645527201645,1.0183471930668224)-- (25.992645527201645,1.0183471930668224);
\draw [line width=0.5pt] (25.992645527201645,1.0183471930668224)-- (25.992645527201645,2.2183471930668226);
\draw [line width=0.5pt] (25.992645527201645,2.2183471930668226)-- (24.702232277835563,2.2106277501362612);
\draw [line width=0.5pt] (16,-1)-- (16,-4);
\draw [line width=0.5pt] (16,-4)-- (22,-4);
\draw [line width=0.5pt] (22,-4)-- (22,-1);
\draw [line width=0.5pt] (22,-1)-- (16,-1);
\draw [line width=0.5pt] (24,-2)-- (24,-4);
\draw [line width=0.5pt] (24,-4)-- (26,-4);
\draw [line width=0.5pt] (26,-4)-- (26,-2);
\draw [line width=0.5pt] (26,-2)-- (24,-2);
\draw [-latex] (22,-3) -- (24,-3);
\draw [-latex] (17,9.5) -- (17,8);
\draw [line width=0.5pt] (17,9.5)-- (8,9.5);
\draw [line width=0.5pt] (7,-2)-- (7,-3);
\draw [line width=0.5pt] (7,-3)-- (8,-3);
\draw [line width=0.5pt] (8,-3)-- (8,-2);
\draw [line width=0.5pt] (8,-2)-- (7,-2);
\draw [line width=0.5pt] (4.5,-2)-- (4.5,-3);
\draw [line width=0.5pt] (4.5,-3)-- (5.5,-3);
\draw [line width=0.5pt] (5.5,-3)-- (5.5,-2);
\draw [line width=0.5pt] (5.5,-2)-- (4.5,-2);
\draw [line width=0.5pt,dash pattern=on 2pt off 2pt] (5.000732091904677,9.595874843365474)-- (5,-2);
\draw [line width=0.5pt] (6,7.5)-- (6,0);
\draw [line width=0.5pt] (6,0)-- (10,2);
\draw [line width=0.5pt] (10,2)-- (10,6);
\draw [line width=0.5pt] (10,6)-- (6,7.5);
\draw [-latex] (8,-2.515725970346784) -- (16,-2.5);
\draw [-latex] (5.5,-2.5) -- (7,-2.5);
\draw [-latex] (10,4.5) -- (11,4.5);
\draw [-latex] (10,2.5) -- (11,2.5);
\draw [-latex] (10,4.5) -- (11,4.5);
\draw [-latex] (23.15123586185288,1) -- (19,-1);
\draw [-latex] (20.710992720268464,1) -- (17,-1);
\draw [-latex] (16.5,1) -- (16.5,-1);
\draw [-latex] (25.344174407262102,1.0183471930668233) -- (21.472802689623073,-1);
\draw [-latex] (25.344174407262102,1.0183471930668233) -- (21.472802689623073,-1);
\draw [-latex] (20.5,4) -- (20.5,3);
\draw [-latex] (23,4) -- (23,2.5);
\draw [-latex] (25,4) -- (24.99605467421118,2.2123854391724334);
\draw [-latex] (20.5,4) -- (20.5,3);
\draw [-latex] (23,4) -- (23,2.5);
\draw [-latex] (25,4) -- (24.99605467421118,2.2123854391724334);
\draw [-latex] (16.5,4) -- (16.5,3);
\draw [-latex] (16.5,4) -- (16.5,3);
\begin{scriptsize}
\draw[color=black] (3.4817190431645946,9.551535680641253) node {$Lr$};
\draw[color=black] (3.043569588004485,7.246488546972936) node {$Lr_{u}$};
\draw[color=black] (2.9673696827592484,4.179442360852282) node {$Ref_{du}$};
\draw[color=black] (3.043569588004485,1.0361962694863962) node {$Ref$};

\draw[color=black] (7.463164092228201,-2.5) node {$F_g$};
\draw[color=black] (4.948567219135397,-2.5) node {$g$};
\draw[color=black] (7.4,9.5) node {$F$};
\draw[color=black] (11.51,6.6) node {$T_{1}$};
\draw[color=black] (11.55,4.4) node {$T_{2}$};
\draw[color=black] (11.7,2.4) node {$T_{3}$};
\draw[color=black] (11.51,0.5) node {$S$};

\draw[color=black] (17,1.8) node {$x_{TT}$};
\draw[color=black] (20.8,1.8) node {$T'_3$};
\draw[color=black] (23.217494501680843,1.8) node {$T'_2$};
\draw[color=black] (25.4,1.7) node {$T'_1$};

\draw[color=black] (25,-3) node {$Sr$};

\draw[color=black] (21,5.9) node {$CSFI$};
\draw[color=black] (19,-2.5) node {$GFE$};
\draw[color=black] (7.787013689520456,3.8) node {$\begin{matrix}
DFE\\
+\\
ST
\end{matrix}$};
\end{scriptsize}
\end{tikzpicture}
    \caption{\footnotesize Efficient Texture Matching with Attention Scheme: We input the low-resolution image, reference down-upsampled, and the reference image, pass it through the Deep Feature Extractor (DFE), and obtain the $k$ relevance texture and score matrices at multiple levels with the SearchAndTransfer (ST). Then we merge the simple features $F$ with the attention textures using a Cross Scale Feature Integration (CSFI). Finally, we refine the partial super-resolution result $x_{TT}$ adding gradient features $F_g$ extracted from the Gradient Density map $g$ to obtain the final Super-Resolution image.}
    \label{fig:main}
\end{figure}
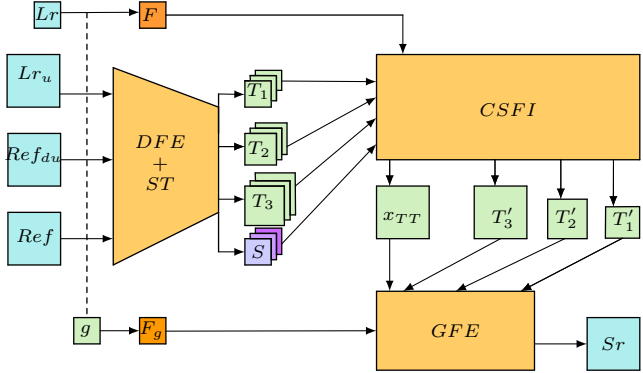

In this section, we proposed Efficient Texture Matching with Attention and Gradient Enhancing for Image Super Resolution (EXTRACTER). It consists of four modules: Deep Feature Extractor (DFE), Search and Transfer Module (STM), Cross-Scale Feature Integration (CSFI), and Gradient Density Enhancing Module (GDE). The main scheme is shown in Fig. \ref{fig:main}.


The model produces a  $4\times $ super-resolution image. It inputs $(Lr_u, Ref_{du}, Ref)$. $Lr_u$ represents the bicubic upsampled low-resolution image and $Ref_{d u}$ represents a bicubic downsampling-upsampling concatenation operation over $Ref$ image. We produce $Q, K, V$ feature maps and find the correlation matrix ($R$) using $Q, K$ normalized inner product between patches. Then we filtered the best patches based on correlation matrix $R$ and then we take the $top_k$ matches for each patch. We integrate the obtained features at three different scales using a Cross Scale Feature Integration\cite{TTSR} and finally, we add gradient density from the LR image to improve structural information and create the Super-Resolution image.

\vspace{-3mm}
\subsection{Deep Feature Extractor}
\vspace{-1mm}

We transform the data into a new representation with more evident complex characteristics at different resolutions. For this, we use the VGG19 \cite{vgg19} backbone (previously trained with ImageNet\cite{imagenet}). Let $(Lr_u, Ref_{du}, Ref)$ be input to our Deep Feature Extractor(DFE). The output of DFE can be formulated as
\begin{eqnarray}\label{DFE}
    Q_i & = & DFE_i(Lr_u) \\
    K_i & = & DFE_i(Ref_{du}) \\
    V_i & = & DFE_i(Ref)
\end{eqnarray}
where $i$ denotes the feature level of the $DFE$. We take three scales of features from VGG19 with output channels $[64, 128, 256]$ and reduce the image $2\times$ to the original scales at each level.

\begin{figure}
    \centering
    \definecolor{ccwwff}{rgb}{0.8,0.4,1}
    \definecolor{cczzff}{rgb}{0.8,0.6,1}
    \definecolor{ccccff}{rgb}{0.8,0.8,1}
    \definecolor{ffzztt}{rgb}{1,0.6,0.2}
    \definecolor{dqfqcq}{rgb}{0.8156862745098039,0.9411764705882353,0.7529411764705882}
    \definecolor{ffccww}{rgb}{1,0.8,0.4}
    \definecolor{afeeee}{rgb}{0.6862745098039216,0.9333333333333333,0.9333333333333333}
    \begin{tikzpicture}[line cap=round,line join=round,>=triangle 45,x=0.38cm,y=0.38cm]
\clip(1.8,-0.5) rectangle (24,10.4);
\fill[line width=0.5pt,color=afeeee,fill=afeeee,fill opacity=1] (2,2) -- (2,0) -- (4,0) -- (4,2) -- cycle;
\fill[line width=0.5pt,color=afeeee,fill=afeeee,fill opacity=1] (2,3) -- (4,3) -- (4,5) -- (2,5) -- cycle;
\fill[line width=0.5pt,color=afeeee,fill=afeeee,fill opacity=1] (1.9679909899501222,8) -- (1.9679909899501222,6) -- (3.9679909899501222,6) -- (3.9679909899501222,8) -- cycle;
\fill[line width=0.5pt,color=ffccww,fill=ffccww,fill opacity=1] (6,7) -- (6,1) -- (7,1) -- (7,7) -- cycle;
\fill[line width=0.5pt,color=afeeee,fill=afeeee,fill opacity=1] (3,10) -- (3,9) -- (4,9) -- (4,10) -- cycle;
\fill[line width=0.5pt,color=dqfqcq,fill=dqfqcq,fill opacity=1] (8,7) -- (8,6) -- (9,6) -- (9,7) -- cycle;
\fill[line width=0.5pt,color=dqfqcq,fill=dqfqcq,fill opacity=1] (8.009344310751096,4.55752184244273) -- (8.009344310751096,3.55752184244273) -- (9.009344310751096,3.55752184244273) -- (9.009344310751096,4.55752184244273) -- cycle;
\fill[line width=0.5pt,color=dqfqcq,fill=dqfqcq,fill opacity=1] (8,2) -- (8,1) -- (9,1) -- (9,2) -- cycle;
\fill[line width=0.5pt,color=ffccww,fill=ffccww,fill opacity=1] (11,6) -- (11,3.5) -- (12,3.5) -- (12,6) -- cycle;
\fill[line width=0.5pt,color=ffccww,fill=ffccww,fill opacity=1] (13.550510806955272,7.003777081049243) -- (13.581866392634966,0.0125149172033205) -- (14.692150404938786,0.03404636874860563) -- (14.675585485912872,7.003387842930539) -- cycle;
\fill[line width=0.5pt,color=ffzztt,fill=ffzztt,fill opacity=1] (8,10) -- (8,9) -- (9,9) -- (9,10) -- cycle;
\fill[line width=0.5pt,color=dqfqcq,fill=dqfqcq,fill opacity=1] (16,6) -- (16,5) -- (17,5) -- (17,6) -- cycle;
\fill[line width=0.5pt,color=ffccww,fill=ffccww,fill opacity=1] (19,7) -- (19,1) -- (20,1) -- (20,7) -- cycle;
\fill[line width=0.5pt,color=ccccff,fill=ccccff,fill opacity=1] (16,3) -- (16,2) -- (17,2) -- (17,3) -- cycle;
\fill[line width=0.5pt,color=dqfqcq,fill=dqfqcq,fill opacity=1] (16.2,6) -- (16.2,6.2) -- (17.2,6.2) -- (17.2,5.2) -- (17,5.2) -- (17,6) -- cycle;
\fill[line width=0.5pt,color=dqfqcq,fill=dqfqcq,fill opacity=0.58] (16.4,6.2) -- (16.4,6.4) -- (17.4,6.4) -- (17.4,5.4) -- (17.2,5.4) -- (17.2,6.2) -- cycle;
\fill[line width=0.5pt,color=ffzztt,fill=ffzztt,fill opacity=1] (16.6,6.4) -- (16.6,6.6) -- (17.6,6.6) -- (17.6,5.6) -- (17.4,5.6) -- (17.4,6.4) -- cycle;
\fill[line width=0.5pt,color=cczzff,fill=cczzff,fill opacity=1] (16.2,3) -- (16.2,3.2) -- (17.2,3.2) -- (17.2,2.2) -- (17,2.2) -- (17,3) -- cycle;
\fill[line width=0.5pt,color=ccwwff,fill=ccwwff,fill opacity=1] (16.4,3.2) -- (16.4,3.4) -- (17.4,3.4) -- (17.4,2.4) -- (17.2,2.4) -- (17.2,3.2) -- cycle;
\fill[line width=0.5pt,color=dqfqcq,fill=dqfqcq,fill opacity=1] (21,10) -- (21,8) -- (23,8) -- (23,10) -- cycle;

\draw [line width=0.5pt] (2,2)-- (2,0);
\draw [line width=0.5pt] (2,0)-- (4,0);
\draw [line width=0.5pt] (4,0)-- (4,2);
\draw [line width=0.5pt] (4,2)-- (2,2);
\draw [line width=0.5pt] (2,3)-- (4,3);
\draw [line width=0.5pt] (4,3)-- (4,5);
\draw [line width=0.5pt] (4,5)-- (2,5);
\draw [line width=0.5pt] (2,5)-- (2,3);
\draw [line width=0.5pt] (1.9679909899501222,8)-- (1.9679909899501222,6);
\draw [line width=0.5pt] (1.9679909899501222,6)-- (3.9679909899501222,6);
\draw [line width=0.5pt] (3.9679909899501222,6)-- (3.9679909899501222,8);
\draw [line width=0.5pt] (3.9679909899501222,8)-- (1.9679909899501222,8);
\draw [line width=0.5pt] (6,7)-- (6,1);
\draw [line width=0.5pt] (6,1)-- (7,1);
\draw [line width=0.5pt] (7,1)-- (7,7);
\draw [line width=0.5pt] (7,7)-- (6,7);
\draw [line width=0.5pt] (3,10)-- (3,9);
\draw [line width=0.5pt] (3,9)-- (4,9);
\draw [line width=0.5pt] (4,9)-- (4,10);
\draw [line width=0.5pt] (4,10)-- (3,10);
\draw [line width=0.5pt] (8,7)-- (8,6);
\draw [line width=0.5pt] (8,6)-- (9,6);
\draw [line width=0.5pt] (9,6)-- (9,7);
\draw [line width=0.5pt] (9,7)-- (8,7);
\draw [line width=0.5pt] (8.009344310751096,4.55752184244273)-- (8.009344310751096,3.55752184244273);
\draw [line width=0.5pt] (8.009344310751096,3.55752184244273)-- (9.009344310751096,3.55752184244273);
\draw [line width=0.5pt] (9.009344310751096,3.55752184244273)-- (9.009344310751096,4.55752184244273);
\draw [line width=0.5pt] (9.009344310751096,4.55752184244273)-- (8.009344310751096,4.55752184244273);
\draw [line width=0.5pt] (8,2)-- (8,1);
\draw [line width=0.5pt] (8,1)-- (9,1);
\draw [line width=0.5pt] (9,1)-- (9,2);
\draw [line width=0.5pt] (9,2)-- (8,2);
\draw [-latex]  (4,1.3853614692292902) -- (6,1.4018479228383658);
\draw [-latex]  (4,4) -- (6,4);
\draw [-latex]  (7,4) -- (8.009344310751096,4);
\draw [-latex]  (7,1.3776638211957717) -- (8,1.3614820161119998);
\draw [line width=0.5pt] (10.007619141753215,5.26464748264292) circle (0.14cm);
\draw [-latex]  (3.9679909899501222,6.5) -- (6,6.5);
\draw [-latex]  (7,6.5) -- (8,6.5);
\draw [-latex]  (10,6.5) -- (9.990022894527401,5.5399080688846984);
\draw [-latex]  (10,4) -- (10,5);
\draw [line width=0.5pt] (9,6.5)-- (10,6.5);
\draw [line width=0.5pt] (9,4)-- (10,4);
\draw [line width=0.5pt] (9.807045568283929,5.453983849673556)-- (10.195428174052571,5.062643062786871);
\draw [line width=0.5pt] (9.813019136949828,5.069176624494906)-- (10.201377366620738,5.460952776893446);
\draw [-latex]  (10.283242779275579,5.25417711046521) -- (10.97327297207902,5.246020546853121);
\draw [line width=0.5pt] (11,6)-- (11,3.5);
\draw [line width=0.5pt] (11,3.5)-- (12,3.5);
\draw [line width=0.5pt] (12,3.5)-- (12,6);
\draw [line width=0.5pt] (12,6)-- (11,6);
\draw [line width=0.5pt] (12.738983436346828,5.254113115800998) circle (0.14cm);
\draw [-latex]  (10,4) -- (11,4);
\draw [line width=0.5pt] (10,6.5)-- (12.720631206149662,6.479026475654235);
\draw [line width=0.5pt] (12,4)-- (12.729892334158812,3.98778304119262);
\draw [-latex]  (12.720631206149662,6.479026475654235) -- (12.721008461282642,5.519559162883495);
\draw [-latex]  (12.729892334158812,3.98778304119262) -- (12.742586070107798,4.988083559792148);
\draw [line width=0.5pt] (12.536146455555418,5.426281239632637)-- (12.931700435238403,5.070687757743541);
\draw [line width=0.5pt] (12.540754165190599,5.076659291674406)-- (12.940515934594133,5.427806396587796);
\draw [line width=0.5pt] (13.550510806955272,7.003777081049243)-- (13.581866392634966,0.0125149172033205);
\draw [line width=0.5pt] (13.581866392634966,0.0125149172033205)-- (14.692150404938786,0.03404636874860563);
\draw [line width=0.5pt] (14.692150404938786,0.03404636874860563)-- (14.675585485912872,7.003387842930539);
\draw [line width=0.5pt] (14.675585485912872,7.003387842930539)-- (13.550510806955272,7.003777081049243);
\draw [-latex]  (13.003182439216808,5.222750869025779) -- (13.558835551453553,5.230264443401585);
\draw [line width=0.5pt] (8,10)-- (8,9);
\draw [line width=0.5pt] (8,9)-- (9,9);
\draw [line width=0.5pt] (9,9)-- (9,10);
\draw [line width=0.5pt] (9,10)-- (8,10);
\draw [line width=0.5pt] (16,6)-- (16,5);
\draw [line width=0.5pt] (16,5)-- (17,5);
\draw [line width=0.5pt] (17,5)-- (17,6);
\draw [line width=0.5pt] (17,6)-- (16,6);
\draw [line width=0.5pt] (19,7)-- (19,1);
\draw [line width=0.5pt] (19,1)-- (20,1);
\draw [line width=0.5pt] (20,1)-- (20,7);
\draw [line width=0.5pt] (20,7)-- (19,7);
\draw [-latex]  (4,9.590958269541302) -- (8,9.60569720076159);
\draw [line width=0.5pt] (9,9.579038744887479)-- (16.545847371316803,9.61397199828316);
\draw [line width=0.5pt] (16,3)-- (16,2);
\draw [line width=0.5pt] (16,2)-- (17,2);
\draw [line width=0.5pt] (17,2)-- (17,3);
\draw [line width=0.5pt] (17,3)-- (16,3);
\draw [line width=0.5pt] (16.2,6)-- (16.2,6.2);
\draw [line width=0.5pt] (16.2,6.2)-- (17.2,6.2);
\draw [line width=0.5pt] (17.2,6.2)-- (17.2,5.2);
\draw [line width=0.5pt] (17.2,5.2)-- (17,5.2);
\draw [line width=0.5pt,color=dqfqcq] (17,5.2)-- (17,6);
\draw [line width=0.5pt,color=dqfqcq] (17,6)-- (16.2,6);
\draw [line width=0.5pt] (16.4,6.2)-- (16.4,6.4);
\draw [line width=0.5pt] (16.4,6.4)-- (17.4,6.4);
\draw [line width=0.5pt] (17.4,6.4)-- (17.4,5.4);
\draw [line width=0.5pt] (17.4,5.4)-- (17.2,5.4);
\draw [line width=0.5pt,color=dqfqcq] (17.2,5.4)-- (17.2,6.2);
\draw [line width=0.5pt,color=dqfqcq] (17.2,6.2)-- (16.4,6.2);
\draw [line width=0.5pt] (16.6,6.4)-- (16.6,6.6);
\draw [line width=0.5pt] (16.6,6.6)-- (17.6,6.6);
\draw [line width=0.5pt] (17.6,6.6)-- (17.6,5.6);
\draw [line width=0.5pt] (17.6,5.6)-- (17.4,5.6);
\draw [line width=0.5pt,color=ffzztt] (17.4,5.6)-- (17.4,6.4);
\draw [line width=0.5pt] (16.2,3)-- (16.2,3.2);
\draw [line width=0.5pt] (17.2,3.2)-- (17.2,2.2);
\draw [line width=0.5pt] (17.2,2.2)-- (17,2.2);
\draw [line width=0.5pt,color=cczzff] (17,2.2)-- (17,3);
\draw [line width=0.5pt,color=cczzff] (17,3)-- (16.2,3);
\draw [line width=0.5pt] (16.4,3.2)-- (16.4,3.4);
\draw [line width=0.5pt] (16.4,3.4)-- (17.4,3.4);
\draw [line width=0.5pt] (17.4,3.4)-- (17.4,2.4);
\draw [line width=0.5pt] (17.4,2.4)-- (17.2,2.4);
\draw [line width=0.5pt,color=ccwwff] (17.2,2.4)-- (17.2,3.2);
\draw [line width=0.5pt,color=ccwwff] (17.2,3.2)-- (16.4,3.2);
\draw [-latex]  (16.569792453239756,9.623751761614729) -- (16.54790000824067,6.828816283398392);
\draw [-latex]  (17.6,5.996341783787673) -- (19,6);
\draw [-latex]  (14.678046330981513,5.968039008953021) -- (16,6);
\draw [-latex]  (14.686278798160346,2.504401495571614) -- (16,2.5);
\draw [-latex]  (17.5,2.5) -- (19,2.5);
\draw [line width=0.5pt] (19.504965063811493,9.586718557489641) circle (0.15cm);
\draw [line width=0.5pt] (16.542846040381004,9.61395810371984)-- (19.815496363564925,9.586718557489641);
\draw [line width=0.5pt] (19.504965063811493,9.897249857243073)-- (19.497573556238088,9.276275239406697);
\draw [-latex]  (19.5,7) -- (19.497573556238088,9.276275239406697);
\draw [line width=0.5pt] (21,10)-- (21,8);
\draw [line width=0.5pt] (21,8)-- (23,8);
\draw [line width=0.5pt] (23,8)-- (23,10);
\draw [line width=0.5pt] (23,10)-- (21,10);
\draw [-latex]  (19.815496363564925,9.586718557489641) -- (20.950630164650512,9.575697283122542);
\draw [-latex]  (9,1.3502350907620073) -- (13.514499981127111,1.3391805850767697);
\draw [line width=0.5pt,dash pattern=on 2pt off 2pt] (5.033187759439004,10.3)-- (5.033187759439004,-0.3);
\draw [line width=0.5pt,dash pattern=on 2pt off 2pt] (5.033187759439004,-0.3)-- (20.3,-0.3);
\draw [line width=0.5pt,dash pattern=on 2pt off 2pt] (20.3,-0.3)-- (20.3,10.3);
\draw [line width=0.5pt,dash pattern=on 2pt off 2pt] (20.3,10.3)-- (5.033187759439004,10.3);
\begin{scriptsize}
\draw[color=black] (2.97082580619384,1.15) node {$Ref$};
\draw[color=black] (3.0039426966857494,4.2177520131295525) node {$Ref_{du}$};
\draw[color=black] (3.0205011419317036,7.264505938385045) node {$Lr_u$};
\draw[color=black] (3.5,9.5) node {$Lr$};

\draw[color=black] (6.514333088828092,4.2) node[rotate=90] {$DFE$};
\draw[color=black] (11.498425107860383,4.8) node[rotate=90] {$Search$};
\draw[color=black] (14.114659456721185,3.605089539029264) node[rotate=90] {$Transfer$};
\draw[color=black] (19.479595716410426,4.2) node[rotate=90] {$Feature Merge$};

\draw[color=black] (8.418554292112855,6.502817457071172) node {$Q$};
\draw[color=black] (8.501346518342627,4.068726005915969) node {$K$};
\draw[color=black] (8.418554292112855,1.5) node {$V$};

\draw[color=black] (8.5,9.5) node {$F$};
\draw[color=black] (16.416283345908855,2.6) node {$S$};
\draw[color=black] (16.48251712689267,5.6914536400194375) node {$T$};
\draw[color=black] (22.013037839041456,9.185285586915683) node {$T_{out}$};
\end{scriptsize}
\end{tikzpicture}
    \caption{\footnotesize Feature extraction and texture search: The model inputs $Lr_u, Ref_{du}, Ref$, pass it through a Deep Feature Extractor (DFE) to perform patch Correlation Search. We use the result as index to select the best $k$-textures by Transfer. Finally, with textures $T$ and soft-attention matrices $S$, we merge them with simple features $F$ from $Lr$ to create $T_{out}$.}
    \label{fig:searchtransfer}
\end{figure}
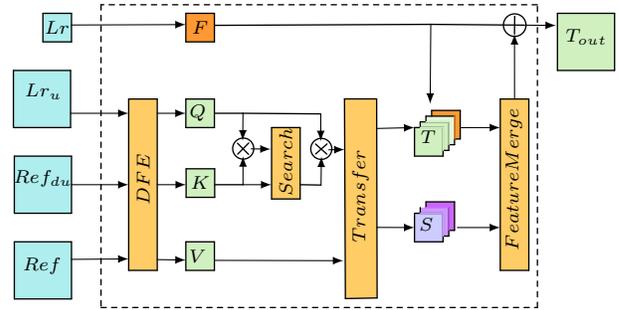

\subsection{Search and Transfer Module}
Let is omit the $i$ index from (\ref{DFE}) for notation simplification. The following calculations are made for a single level of DFE, is is depict at Fig. \ref{fig:searchtransfer}. We infer correlations between $LR_u$ and $ Ref_{du}$ using attention via $Q$ and $K$ at two stages. First we divide $Q, K$ into overlaping patches $q_i : i  \in  [1,2,\ldots, H_{LR} \times W_{LR} / s^2]$ and $k_j : j  \in  [1,2,\ldots, H_{Ref} \times W_{Ref}/s^2 ]$, respectively, where $s$ is the stride setput for patch displacement. In experiments, we use a window of $6$ and stride $s = 2$. The correlation matrix is computed as the normalized inner product
\begin{equation}
    \label{eq:scores}
	c'_{i,j} = \left\langle \frac{q_i} {|| q_i ||}, \frac{k_j}{|| k_j ||}\right\rangle .
\end{equation}
Next, we keep the best score indices of the $k_j$ patches for each of $q_i$  
$H' = \arg\max_j (C')$ . Using the $H'$ matrix as index, we extract the most relevance patches of $K$ as
$ K' = K_{H'}$.
Following, we  use a re-search strategy by keeping the best score indices of the $k'_j$ normalized patches for each of $q_i$  using the $top_u$ largest matches
\begin{equation}
    \label{eq:top_k}
	H, S = top_u (C) \textup{  with  } c_{i,j} = \left\langle \frac{q_i}{|| q_i ||}, \frac{k'_j}{|| k'_j ||}\right\rangle 
\end{equation}
with $S, H$ tensors containing the $u$-maximum scores and index for $C$; i.e.,
\begin{equation}
    H_0  =  \arg \max_j C_{i j} \textup{ ,}\hspace{2 mm} S_0  =  \max_j C_{i j}
\end{equation}
and $H_1, S_1$ be the second maximum indices and scores matches, etc. Now, we select the best textures from $V$ using the $H_i$, $i = 1, \dots, u$ matrices:
$
    T_i = V_{H_i}
$.
So that, we extract the best matches using the hard attention matrix as index. Finally, for an output of the Initial Feature Extractor (IFE) of LR image, denote as
$F = IFE(x)$.
Hence, we integrate the found features $T_i$:
\begin{equation}
    F_{TT} = F + \sum_{i=1}^u Conv_i (Concat(F, T_i \otimes S_i)) \otimes S_i;
\end{equation}
where $\otimes$, $Conv_i(\cdot)$ and $Contat(\cdot)$ denotes  element-wise multiplication, convolutional $3 \times 3$ and concatenation blocks, respectively. 

\subsection{Cross-Scale Feature Integration}
Inspired by SoTA methods for style/texture transferring \cite{gatys2016neural,zeng2019learning,TTSR}, We integrate the previous attention results at different  scales following \cite{TTSR}; this can be modeled as
\begin{equation*}
    \label{eq:csint}
    x_{TT}, T_1, T_2, T_3 = CCFI(\{F_{TT}^{(i)}\}_{1=1,2,\ldots});
\end{equation*}
where $x_{TT}$ is the merged super-resolution texture and $T_1$,  $T_2$, $T_3$  are the syntetized textures.

\vspace{-5mm}
\subsection{Gradient Enhancing Density Module}
\vspace{-1mm}

To give more information about the structure of the low-resolution image, some work has been done \cite{grad1, grad2}. We incorporate a Gradient Enhancing module for adding structural and edge information to the partial output of the $CSFI(\cdot)$. First, we extract the Gradient Density for each of the RGB image channels we convolve the Image with $3 \times 3$ Sobel filters kernels \cite{sobel} from $x$ and $y$ derivative directions; $K_x$ and $K_y$, respectively.
\mycomment{
\begin{equation*}
    K_x = \left\lbrack \begin{matrix}
                    -1 & 0 & 1 \\
                    -2 & 0 & 2\\
                    -1 & 0 & 1
    \end{matrix}\right\rbrack, 
    K_y = \left\lbrack \begin{matrix}
                    1 & 2 & 1 \\
                    0 & 0 & 0\\
                    -1 & -2 & -1
    \end{matrix}\right\rbrack
\end{equation*}
}
and calculate Gradient Density as
\begin{equation*} \label{gradient}
    GD(I) = \sqrt{(K_x*I)^2 + (K_y*I)^2}.
\end{equation*}
Now, we pass the image gradient density $g$ through a residual feature extractor:
$
    F_g = GFE(g)
$. Finally, using the output from $CSFI(\cdot)$ : $x_{TT}, T_1, T_2, T_3$, the SR  image is formulate as
\begin{eqnarray*}
    x_{1g} & = & RB_1(Conv(Concat(F_g, T_3))) \\
    x_{2g} & = & RB_2(Conv(Concat(x_{1g}\uparrow, T_2))) \\
    x_{3g} & = & RB_3(Conv(Concat(x_{3g}\uparrow, T_3))) \\
    SR     & = & Conv(Concat(x_{3g}, x_{TT}))
\end{eqnarray*}
where $RB(\cdot)$ represents a residual scheme and $\uparrow$ is $2\times $ bicubic upsampling. 

\subsection{Loss Function}
The overall loss is
\begin{equation}\label{eq:loss}
    \mathcal{L}_{total} = \lambda_1 \mathcal{L}_{rec} +\lambda_2  \mathcal{L}_{perc} + \lambda_3 \mathcal{L}_{grad} + \lambda_4 \mathcal{L}_{adv}
\end{equation}
where
\begin{equation*}
    \mathcal{L}_{rec} = (chw)^{-1}|| SR - HR ||_1 ,
\end{equation*}
with $c, h, w$ the channel, height, weight of the $HR$ image.
In the aim of enhacing the similarity of the feature space representation of the generated image and the $SR$ image using the $vgg19$ feature space \cite{vgg19_fe, vgg19_fe1}, we use
\begin{equation*}
    \mathcal{L}_{perc} = (c_ih_iw_i)^{-1}|| vgg19_i(SR) - vgg19_i(HR) ||_1,
\end{equation*}
with $c_i, h_i, w_i$ the channel, height, weight at the correspoinding $i$ level.
For structural similarity enhacing, we introduce Gradient Density Loss using (\ref{gradient})
\begin{equation*}
    \mathcal{L}_{grad} = (chw)^{-1} || GD(SR) - GD(HR) ||_1,
\end{equation*}
with $c_i, h_i, w_i$ the channel, height, weight at the correspoinding $i$ level. Similar to \cite{TTSR, esteban}, we use a WGAN-GP for more stable training. 
This loss is described as
	\begin{eqnarray*}
		\mathcal{L}_D & = & \mathbb{E}_{\tilde{x} \sim \mathbb{P}_g} \left[ D(\tilde{x}) \right] 
		-\mathbb{E}_{x \sim \mathbb{P}_r} \left[ D(x) \right] \\
		&   & +\lambda\mathbb{E}_{\hat{x} \sim \mathbb{P}_{\hat{x}}} \left[ \left\Vert \nabla_{\hat{x}} D(\hat{x})\right\Vert_{2} - 1)^2 \right], \\
		\mathcal{L}_G & = & - \mathbb{E}_{\tilde{x} \sim \mathbb{P}_g} \left[ D(\tilde{x}) \right].
	\end{eqnarray*}

\vspace{-5mm}
\subsection{Implementation Details}
\vspace{-1mm}

The window size for extracting patches is set as $k = 6$ with padding $p = 2$ and a stride of $s = 2$. In experiments, we explore other configurations. The architecture for the CSFI model is $[16,8,4]$,  $[9 ,9, 9]$ for GDE and $4$ residual blocks for IFE's. For the correlation matrix, we use only the deepest feature extractor level to perform matrix multiplication. We use data augmentation for training by randomly flipping up-down and left-right followed by a random rotation of $90 ^\circ,180^ \circ, 270^ \circ$ with a batch fixed to $9$. The weights of the loss coefficients are $1, 1e^{-2}, 1e^{-3}, 1e^{-3}$ in the same order of equation (\ref{eq:loss}). An Adam optimizer with $lr = 1e^{-4}$, $\beta_1 = 0.9$, $\beta_2 = 0.999$ and default $\epsilon = 1e^{-8}$. All the experiments were performed in a single GPU NVIDIA GeForce RTX 3090 using the pytorch framework.

\vspace{-5mm}
\section{EXPERIMENTS AND RESULTS}
\vspace{-3mm}

Following the recent work, we use two metrics to evaluate the results: Peak Signal to Noise Ratio (PSNR) and Structure Similarity Index (SSIM) \cite{hore2010imagemetric}. 
\mycomment{
The PSNR is defined as
\begin{equation*}
    \label{eq:PSNR}
	PSNR(f, g) = 10 \log_{10} (255/\mathrm{mse}(f, g))
\end{equation*}
where $\mathrm{mse}$ is the Mean Square Error. SSIM is defined as
\begin{equation*}
	SSIM(f, g) = l(f, g) c(f, g) s(f, g)
\end{equation*}
with
\begin{align*}\label{loss}
	l(f, g) &=  (2 \mu_f \mu_g + k_1) / (\mu_f^2 + \mu_g^2 + k_1), \\
	c(f, g) &=  (2 \sigma_f \sigma_g + k_2)/(\sigma_f^2 + \sigma_g^2 + k_2),\\
	s(f, g) &=  (2 \sigma_{fg} + k_3)/(\sigma_f \sigma_g + k_3); 
\end{align*}
where $\,\,\{c_k\}_{k=1,2,3}\,\,$ are parameters (constants) of the metrics, $\,\,\,\,\{\mu_z\}_{z=f, g}\,\,\,$ and $\{\sigma^2_z\}_{z=f,g, fg}$ are the mean values and (co)variances, respectively. The first term compares the luminescence, the second term the contrast and the third term the structure. 

\subsection{Datasets}}
We conduct the training using CUFED5 Dataset \cite{cufed}. It contains 11,871 pairs consisting of an input and reference image. There are 126 testing images, each having 4 reference images with different similarity levels. We also evaluate our method using different text sets such as Sun80 \cite{sun80}, Urban100 \cite{urban100}, and Set14\cite{set14}. Sun80 contains 80 natural images, each of them paired with several reference images. Urban100 and Set14 do not have reference images so we took it randomly from the same dataset. All the SR results are evaluated of PSRN and SSIM on the Y channel of YCbCr space.
Following the SOTA methods, we train our model using the train set from CUFED5 and test it on the CUFED5 test set, Sun80, Urban100, and Set14. Two versions of our model were trained, the first one trained only using reconstruction loss and the second using all losses. EXTRACTER-rec outperforms recent methods despite using a bigger window size, as we can see in Table \ref{table:metrics}. We observe better visual results when all losses were used, Fig. \ref{fig:main} illustrates some visual results with other novel models. 
We study different configurations for our model. Table \ref{table:kernel} shows the number of parameters and the correlation matrix shape during the training phase for the CUFED5 dataset. We found that our method reduces $4\times$ the shape from the attention mechanism. Table \ref{table:kernel} shows the effectiveness of changing the kernel size for the test phase using large image size datasets such as Sun80 and Urban100.

\begin{table}[t]
\footnotesize
  \centering
    \setlength\tabcolsep{4pt} 
  \begin{tabular}{lcccc}
 \hline
 Method          & CUFED5              & Sun80         & Urban100      & Set14\\ 
 \hline\hline
 \mycomment{
 Bicubic         & 24.31 / .688       & 28.65 / .767     & 23.23 / .663     & 26.10 / .720 \\
 SRCNN           & 25.34 / .746       & 28.26 / .781     & 24.42 / .738     & 26.41 / .733 \\
 MDSR            & 25.93 / .7772      & 28.52 / .792     & 25.51 / .783     & -            \\
 RDN             & 25.95 / .769       & 29.63 / .806     & 25.38 / .768     & 27.76 / .770 \\
 RCGAN           & 26.06 / .769       & 29.86 / .810     & 25.43 / .768     & -            \\
 ESRGAN          & 21.93 / .634       & 24.18 / .651     & 20.91 / .620     & -            \\
 RSRGAN          & 22.31 / .635       & 25.60 / .667     & 21.47 / .624     & -            \\
 \hline
 }
 SRNTT           & 25.61 / .764       & 27.59 / .756     & 25.09 / .774     & 26.73 / .731 \\
 SRNTT-rec       & 26.24 / .784       & 28.54 / .793     & 25.50 / .784     & 27.68 / .766 \\
 TTSR            & 25.63 / .765       & 28.59 / .774     & 24.69 / .748     & 26.88 / .748 \\
 TTSR-rec        & 27.03 / .802       & 30.02 / .814     & 25.88 / .784     & \bf  28.10 / .782 \\
 SSEN-rec        & 26.78 / .791       & -                & -                & -            \\
 DPFSR           & 25.23 / .749       & 28.59 / .774     & 24.35 / .734     & -            \\
 DPFSR-rec       & \bf 27.25 / .808   & \bf 30.10 / .815 & \bf 26.03 / .787 & -            \\
 $C^2$- Matching & 27.16 / .805       & 29.75 / .799     & 25.52 / .764     & -            \\
 Extracter           & 26.40 / .789       & 29.02 / .789     & 24.72 / .752     & 26.50/.740   \\
 Extracter-rec       & {\bf 27.29 / .811} & \bf 30.02 / .816 & \bf 26.04 / .785 & \bf 28.09 / .782 \\

 \hline
\end{tabular}
\caption{ \footnotesize Quatitative metrics of the generated images using PSNR / SSIM. The 2-highest scores are denoted in black.}
\label{table:metrics}
\end{table}

\begin{table}[ht!]
 \footnotesize
  \centering
    \setlength\tabcolsep{4pt} 
  \begin{tabular}{lccc}
 \hline
 Method         & Params. (M) & Kernel size & corr. matrix shape \\ 
 \hline\hline
 TTSR           & 6.73        & $3 \times  3$   & $1600 \times  1600$ \\
 DPFSR          & 6.91        & $3 \times 3 $   & $1600 \times  1600$ \\
 Extracter      & 9.31        & $6 \times  6$   & $800  \times  800 $ \\

 \hline
\end{tabular}
\caption{\footnotesize Model parameters and shape of the training correlation matrix. Our method reduce significantly the matrix multiplication cost by extracting larger patches.}
\label{table:params}
\end{table}

\begin{table}[ht!]
 \footnotesize
  \centering
    \setlength\tabcolsep{4pt} 
  \begin{tabular}{lccc}
 \hline
 Kernel Size    & Sun80        & Urban100  \\ 
 \hline\hline
 $3 \times 3$   & OFM          & OFM       \\
 $6 \times 6$   & 30.02 / .816 & 26.04 / .785 \\
 $12 \times 12$ & 29.98 / .814 & 25.74 / .781  \\
 \hline
\end{tabular}
\caption{\footnotesize Kernel size when obtaining patches for PSNR / SSIM metrics with our model. All comparations where made on single GPU. Models using $3 \times  3$ kernel like TTSR and DPFSR produces Out of Memory (OFM) due the large image dimensions on Sun80 and Uban100 datasets.}
\label{table:kernel}
\end{table}

\begin{figure}[ht!]
\begin{minipage}[b]{1.0\linewidth}
  \centering
  \footnotesize{
  \begin{tabular}{ccc}
     Input          & HR              & SRNTT         \\ 
     \hline
     Reference     & TTSR              & EXTRACTER         \\ 
\end{tabular}}
  \centerline{\includegraphics[width=7cm]{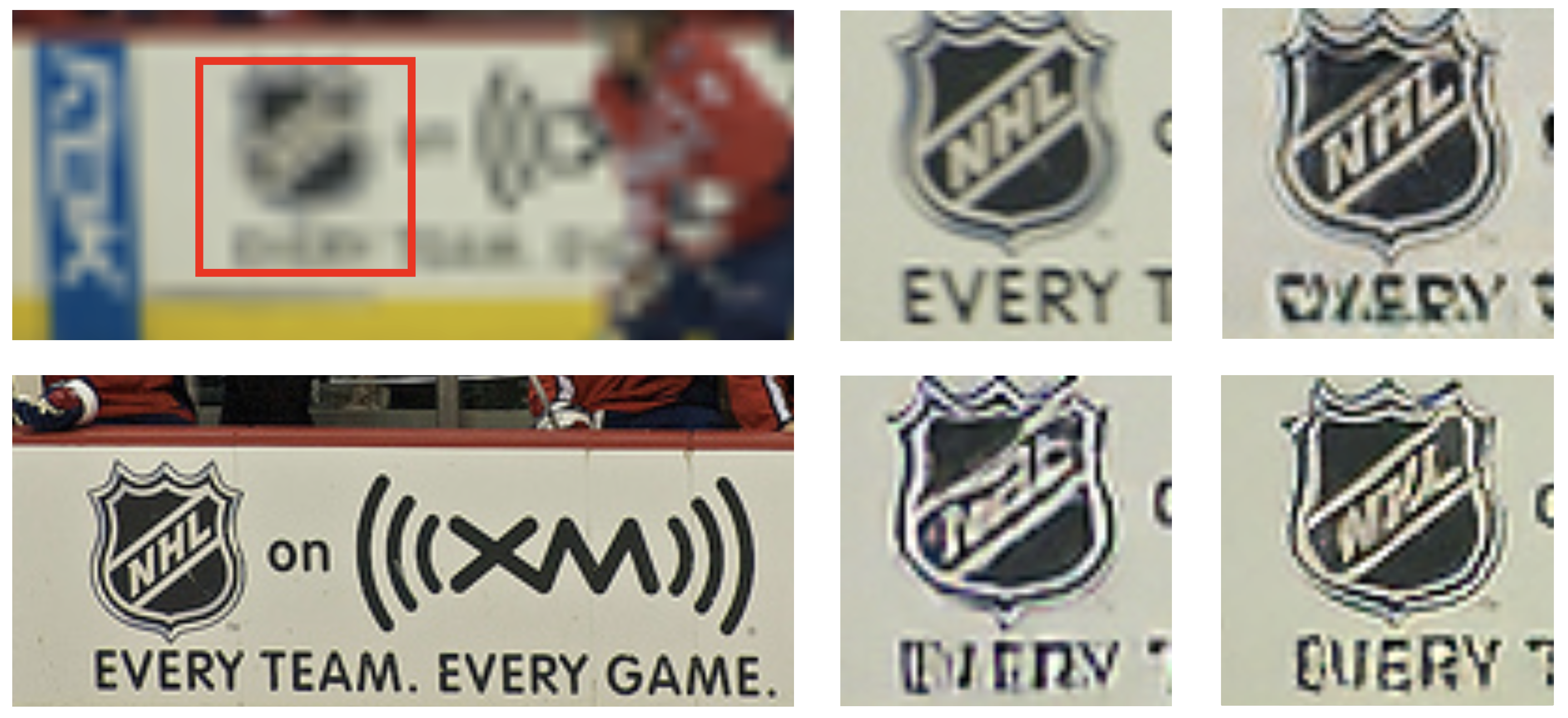}}
\end{minipage}
\begin{minipage}[b]{1.0\linewidth}
  \centering
    \footnotesize{
  \begin{tabular}{ccc}
     Input          & HR              & TTSR         \\ 
     \hline
                    & Reference              & EXTRACTER         \\ 
\end{tabular}}
  \centerline{\includegraphics[width=7cm]{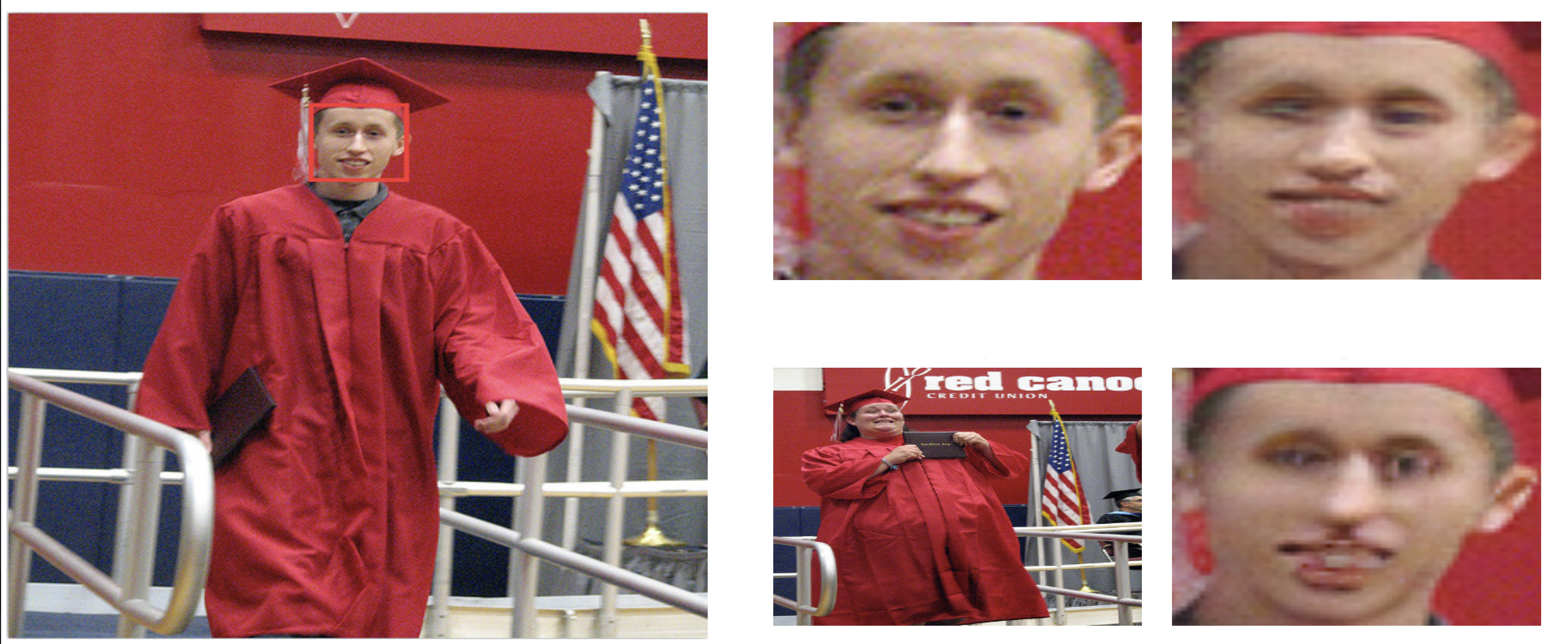}}
\end{minipage}
\caption{Experimental results: we compare our model with available testing models online.}
\label{fig:res}
\end{figure}

\section{CONCLUSIONS AND FUTURE WORK}

In this paper, we propose a novel deep texture search with more efficient memory usage for RefSR. The proposed model consists of a learnable Deep Feature Extractor, a Search and Transfer Module that uses the top-$k$ matches between the Lr and Ref patches for transferring textures in a more efficient memory usage way than SOTA methods by using larger windows, a Cross Scale Feature Integrator and, finally, a Gradient Enhancing Density module. Our experiments demonstrate the competitive performance of EXTRACTER over the recent attention mechanisms for RefSR using PSRN and SSIM metrics. The ablation studies demonstrate the efficiency of managing larger windows when using large-scale images, resulting in a non-out-of-memory as other recent methods. In the future, we would like to enhance our model by changing the CSFI for a simpler network to reduce training time,  using the transferring mechanisms to refine generative models, and exploring RefSR real-world applications, such as satellite super-resolution and movie super-resolution.

\noindent {\bf Acknowledges.} Work supported by Conacyt, Mexico (Grant CB-A1-43858) and E. Reyes Scholarship.
\newpage





\bibliographystyle{IEEEbib}
\bibliography{reyes_extracter_2023}

\end{document}